\documentclass[a4paper,fleqn]{cas-sc}

\usepackage[numbers]{natbib}
\usepackage[linesnumbered,ruled,vlined]{algorithm2e}

\def\tsc#1{\csdef{#1}{\textsc{\lowercase{#1}}\xspace}}
\tsc{WGM}
\tsc{QE}
\tsc{EP}
\tsc{PMS}
\tsc{BEC}
\tsc{DE}

\begin{document}
\let\WriteBookmarks\relax
\def\floatpagepagefraction{1}
\def\textpagefraction{.001}

\shorttitle{Self-supervised Monocular Depth Estimation on Water Scenes via Specular Reflection Prior}

\shortauthors{Zhengyang Lu et~al.}

\title [mode = title]{Self-supervised Monocular Depth Estimation on Water Scenes via Specular Reflection Prior}                      
\tnotemark[1]

\tnotetext[1]{This work is supported by the National Natural Science Foundation of China under Grant 62173160 and Grant 61573168.}

%
\author[1]{Zhengyang Lu}

\affiliation[1]{organization={the Key Laboratory of Advanced Process Control for Light Industry (Ministry of Education), Jiangnan University},
    city={Wuxi},
    country={China}}

\author[1]{Ying Chen}

\cormark[1]

\cortext[cor1]{Corresponding author}

\nonumnote{luzhengyang@jiangnan.edu.cn(Z.Lu); chenying@jiangnan.edu.cn(Y.Chen)}

\begin{abstract}
Monocular depth estimation from a single image is an ill-posed problem for computer vision due to insufficient reliable cues as the prior knowledge.
Besides the inter-frame supervision, namely stereo and adjacent frames, extensive prior information is available in the same frame.
Reflections from specular surfaces, informative intra-frame priors, enable us to reformulate the ill-posed depth estimation task as a multi-view synthesis.
This paper proposes the first self-supervision for deep-learning depth estimation on water scenes via intra-frame priors, known as reflection supervision and geometrical constraints.
In the first stage, a water segmentation network is performed to separate the reflection components from the entire image.
Next, we construct a self-supervised framework to predict the target appearance from reflections, perceived as other perspectives.
The photometric re-projection error, incorporating SmoothL1 and a novel photometric adaptive SSIM, is formulated to optimize pose and depth estimation by  aligning the transformed virtual depths and source ones. 
As a supplement, the water surface is determined from real and virtual camera positions, which complement the depth of the water area.
Furthermore, to alleviate these laborious ground truth annotations, we introduce a large-scale water reflection scene (WRS) dataset rendered from Unreal Engine 4.
Extensive experiments on the WRS dataset prove the feasibility of the proposed method compared to state-of-the-art depth estimation techniques.
\end{abstract}




\maketitle

\section{Introduction}

Depth estimation is a crucial problem in computer vision, which is a basis for scene understanding and could guide other vision tasks such as classification, tracking, segmentation, and detection. 
This long-standing ill-posed problem has been extensively investigated for decades, aiming to restore one depth map from one or more RGB images.
Nevertheless, humans learn from prior information in realistic scenarios, allowing us to hypothesize plausible depth estimates for intricate scenes \citep{hochberg1952familiar}.

\begin{figure}[tbp]
\centering
\includegraphics[width=0.7\linewidth]{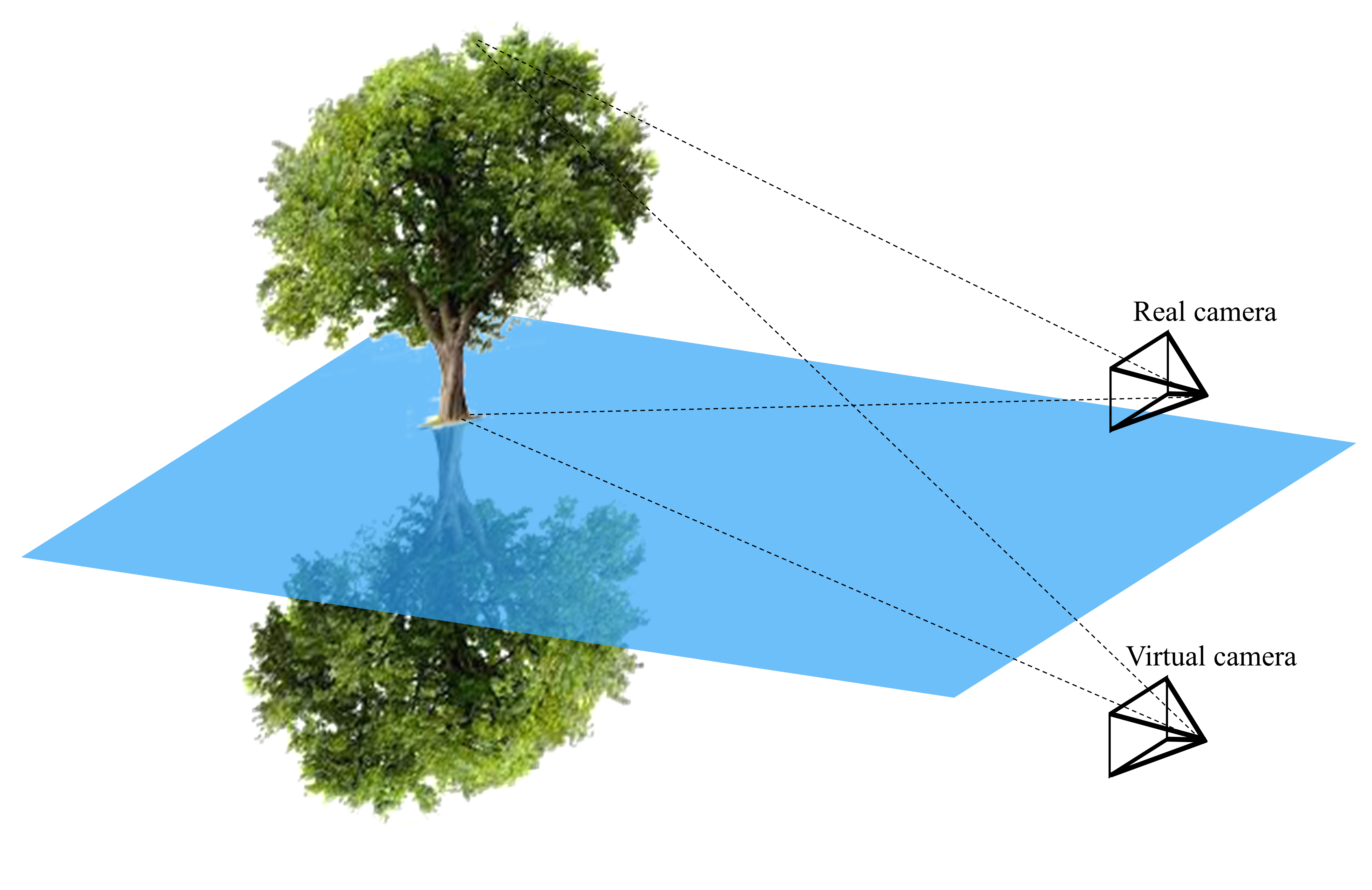}
\caption{Schematic of depth estimation on the reflective scene.
The simultaneous appearance of the inverted and raw image reformulates the ill-posed depth estimation task as an interpretable multi-view synthesis problem.}
\label{fig:refsch}
\end{figure}

\begin{figure*}[tbp]
	\centering
	\includegraphics[width=\linewidth]{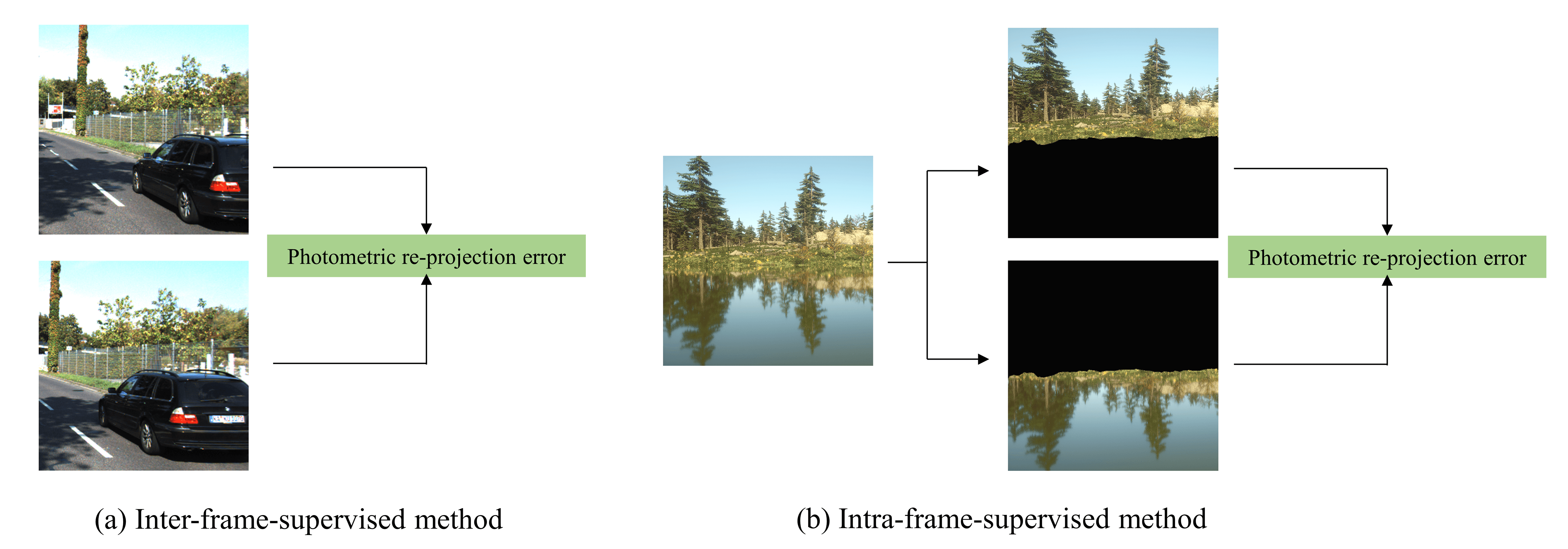}
	\caption{Differences in inter- and intra-frame supervision methods. Reflection information enables self-supervised depth estimation in single frames.}
	\label{fig:interintra}
\end{figure*}

Most existing depth estimation models are optimized in two training manners, namely depth-supervised and self-supervised.
The depth-supervised methods suffers from low interpretability and requires numerous images and ground-truth depth annotations, thus self-supervised methods become the dominant solution, which are supervised by stereo and adjacent frames.
Most self-supervised depth estimation techniques rely on inter-frame supervision mechanisms. These methods successfully eliminate the laborious process of manual annotations, but only exploit limited inter-frame information.
Consequently, extensive information-rich depth features in reflection scenes remain unexploited.
Besides the inter-frame supervision, extensive prior information can be exploited in single frames, which makes it possible to create an intra-frame-supervised depth estimation framework.
Therefore, reflections from the specular surface, which is the intra-frame prior information that is prone to matching, allow us to reformulate the arbitrary end-to-end depth estimation solution as a plausible multi-view synthesis, as shown in Figure \ref{fig:refsch}.
Since specular reflections are the equivalent to viewing the same scenes from another perspective.
As shown in Figure \ref{fig:interintra}, reflective sources in the same frame can be invoked as supervision for the depth model, such as mirrors, car paintwork and water surfaces. 
Such methods can be applied to the depth reconstruction of reflective surfaces and multi-view stereo matching in reflective cases, because the reflection image is regarded as another view.

The first symmetric-scene depth estimator \citep{yang2015depth} infers accurate depths from water reflections, but suffers from the strict assumption that the reflection interface requires to be parallel to the image horizon line.
Yet, current deep-learning technology allows us to construct a pose estimator to solve arbitrary viewpoint issues.

\begin{figure}[tbp]
	\centering
	\includegraphics[width=0.6\linewidth]{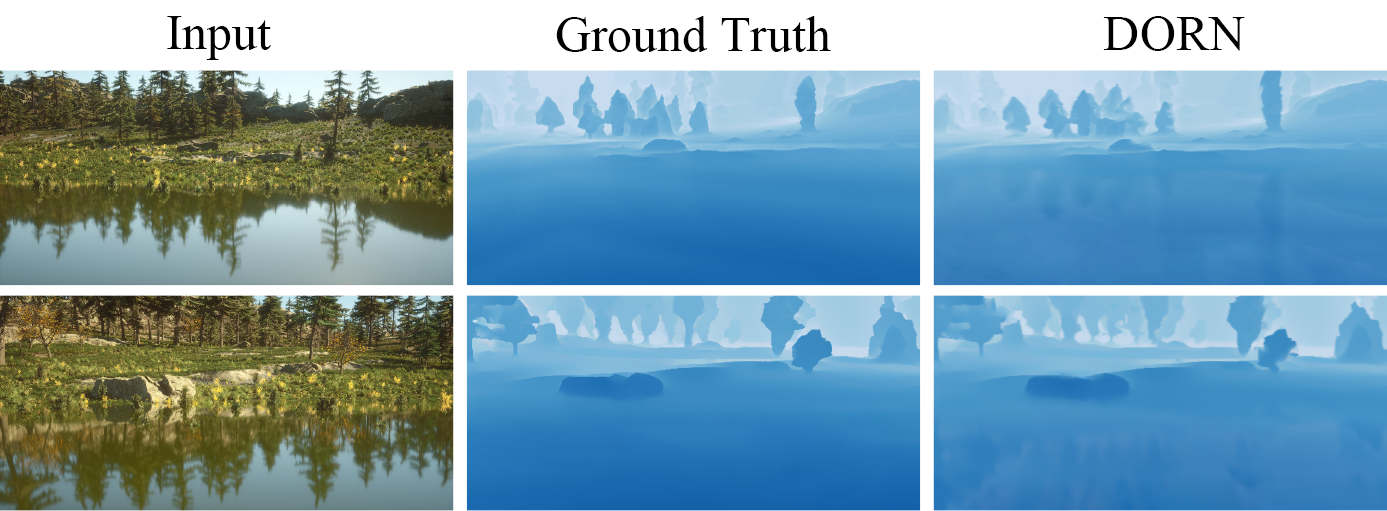}
	\caption{Depth from a single image in water reflection scenes.
		DORN, an end-to-end depth estimation method, produces blurred, deluded results.}
	\label{fig:inidemo}
\end{figure}

Up to now, far too little attention has been paid to the intra-frame-supervised methods.
The problem addressed in this paper is to estimate accurate depth using intra-frame information from a single frame sample in reflective scenes.
The difficulty of reflection-supervised methods involves, on the one hand, separating the reflective priors and, on the other hand, matching the source and virtual patterns which are heavily attenuated due to the light reflection principle.
As shown in Figure \ref{fig:inidemo}, most existing end-to-end methods are deceived by reflections to provide erroneous depth estimates.

This paper presents the first deep-learning work for depth estimation on water scenes via specular reflection priors.
Therefore, we introduce a general depth estimation framework for intra-frame information interaction in the presence of specular reflections.
The pipeline of the proposed work can be implemented in two steps: end-to-end water segmentation and self-supervised depth estimation. 
For the water segmentation, there is an end-to-end U-Net\citep{ronneberger2015u} backbone with Dice loss \citep{milletari2016v}. 
In contrast to the widely-used depth-supervised depth estimation, the photometric re-projection supervision, which estimates the pose of the source and corresponding virtual camera, is a solution to extract this interpretable depth from the model.
The classical photometric re-projection error, which combines SSIM \citep{wang2004image} and SmoothL1 \citep{girshick2015fast}, has scene limitations in matching source and reflected images due to the luminosity attenuation.
Since the luminance and contrast of the reflected images decrease proportionally, Photometric Adaptive SSIM (PASSIM), which eliminates the luminance component and emphasizes the contrast and structure comparison, is proposed as an alternative to SSIM.
To address the absence of the dataset on reflection scenes, we create a large-scale specular reflection dataset, the Water Reflection Scene dataset (WRS), containing 3251 RGB images and depth pairs rendered from Unreal Engine 4.
Furthermore, experimental results demonstrate that our framework outperforms the most advanced models.

The major contributions of the proposed depth estimation method in water reflection scenes can be summarized as follows:
\begin{itemize}
\item{We present an intra-frame-supervised depth estimation via specular reflection, the first deep-learning work to reformulate single-frame reflection image depth reconstruction as a multi-view synthesis problem. To construct a general framework that is prone to re-implement, most lightweight backbones are employed, standard U-Nets for water segmentation and depth estimation, and the ResNet18 for pose estimation.}
\item{To match the reflections and source patterns, the photometric adaptive SSIM, which is developed from SSIM, is introduced to focus on local contrast and structure.}
\item{We addressed the absence of reflection scene datasets by creating the Water Reflection Scene (WRS) dataset. It includes 3251 RGB images and depth maps rendered from Unreal Engine 4, serving as a valuable resource for reflection analysis and related computer vision research.}
\item{Our experimental results demonstrate the proposed monocular depth estimation on water scenes is superior to the previous depth estimation methods and show the state-of-the-art performance of the WRS dataset. Furthermore, we trained the proposed model with images collected from the Internet and provided accurate visualizations to validate its feasibility.}

\end{itemize}

This paper is structured as follows.
Section II reviews related works on deep-learning monocular depth estimation.
Section III describes the monocular depth estimation on water scenes via specular reflection, 
and Section IV assesses the proposed methodology through extensive experiments and discusses limitations.
Finally, Section V concludes the paper and predicts further research trends.

\section{Related Works}

In this section, we provide a brief structured review of recent developments in monocular depth estimation using deep learning techniques.
These methods estimate one depth map from one or more RGB images in 3D scenes, captured using cameras with known or unknown intrinsic and extrinsic properties.
During the early period, depth maps were estimated depending on hand-made cues, such as shadow \citep{zhang1999shape}, focus \citep{tang2015depth}, and keypoint \citep{lowe1999object}.
However, in the last decade, deep-learning methods have provided unparalleled accuracy in depth estimation by replacing hand-designed feature extraction and model fitting with data-driven techniques.
The following section introduces and classifies deep-learning models from the aspect of training manners: depth-supervised and self-supervised methods for monocular depth estimation.
The purpose of investigating these advanced monocular depth estimation methods is to locate problems, which have limited accuracy and interpretability  until now.

\subsection{Depth-supervised Learning Methods}

Depth-supervised learning methods aim to minimize the errors between the predictions and ground-truth depths by various loss constraints.

CNN-based depth estimation can be categorized as two implementations: direct and relative depth estimation for each pixel.
For direct depth estimation, the first depth-supervised work was proposed by Eigen \textit{et~al.} \citep{eigen2014depth}.
Eigen \textit{et~al.} \citep{eigen2014depth} developed a coarse-to-fine framework, with the coarse network estimating the rough depth, and the fine network learning to refine the vague depth map through a multi-layer structure.

On the basis of VGG-16 \citep{simonyan2014very}, Li \textit{et~al.} \citep{li2017two} demonstrated a multi-stream model: Li \textit{et~al.} \citep{li2017two} proposed a two-streamed system based on VGG-16 \citep{simonyan2014very}: one stream for depth prediction and another for depth gradients, combined via a fusion block to create a continuous depth. The end-to-end system was supervised by the depth and gradient, enhancing each stream's generalization mutually. 
Laina \textit{et~al.} \citep{laina2016deeper} illustrated a Fully Convolutional Residual Network (FCRN) to fit the mapping between monocular images and depths with the reverse Huber loss, which is driven by value distributions.
A plug-and-play convolutional neural field model \citep{liu2015learning} for predicting depths from monocular images was developed with the intention of mutually exploring the potential of deep network and continuous Conditional Random Fields (CRFs).
In the same work, they also presented an efficacious model based on fully convolutional networks and a super-pixel pooling approach to speed up the patch-wise convolutions. 
Ricci \textit{et~al.} \citep{xu2017multi} came up with a cascade of multiple CRFs and a aggregated graphical model. They also proved that it is a feasible approach to fuse multi-scale features with multi-level CRF integration cascades.
For bootstrapping densely encoded information to depth predictions, Lee \textit{et~al.} \citep{lee2019big} provided a solution called Big To Small (BTS) in 2019, which exploits local planar guidance layers located at multiple stages in the decoding process.
In 2018, Fu \textit{et~al.} \citep{fu2018deep} proposed a Deep Ordinal Regression Network (DORN), which applied a spacing-increasing discretization strategy to discretize depth, to alleviate the problem of sluggish convergence and mismatched local features.
In depth estimation using Adaptive Bins (AdaBins), bhat \textit{et~al.} \citep{bhat2021adabins} propose a transformer-based architecture block to improve entire depth estimation through global information processing.
To avoid failing to utilize underlying properties of well-encoded features, Song \textit{et~al.} \citep{song2021monocular} incorporates the Laplacian pyramid (LapPy) into the decoder architecture in 2021.
To improve the robustness of depth estimation models, Lu \textit{et~al.} \citep{lu2022pyramid} introduced a Pyramid Frequency Network with spatial attention Residual Refinement module (PFN). The frequency division strategy was designed to extract features from multiple frequency bands.
Utilizing a multi-head attention mechanism, Yuan \textit{et~al.} \citep{yuan2022new} enhanced inter-node relationships and generated an optimized depth map. They constructed a bottom-up-top-down architecture with a neural window FC-CRFs module (NeWCRFs) as the decoder and a vision transformer as the encoder.

For relative depth estimation, Zoran \textit{et~al.} \citep{zoran2015learning} constructed a model for inferring depth information based on the relative relationship between point pairs in the same image. They enhance the robustness of the relative depth estimation method and provide qualitatively different information because ordinal relationships are invariant to monotonic transformations.
Following the preliminary relative depth estimation method, Chen \textit{et~al.} \citep{chen2016single} presented a multi-scale model that estimated pixel-level depth through fitting relative depth mappings.
The model reconstructed the depths from single images in an unconstrained scene, and was supervised by a relative depth loss function.

Generative Adversarial Networks (GANs) \citep{goodfellow2014generative} were a tricky deep-learning strategy that can generate depth maps close to ground truth.
Jung \textit{et~al.} \citep{jung2017depth} first adapted GANs to monocular depth estimation, using a global network extracting global features and a refinement network estimating local features from the input image.
Lu \textit{et~al.} \citep{lu2021ga} demonstrate a high-order convolutional spatial propagation dense-connected U-Net for reducing information transmission loss,
which employs a discriminator with a modified binary-relationship loss function.

Depth-supervised methods have been extensively investigated and applied in monocular depth estimation, primarily including CNN-based and GAN-based models, where the CNN mainly learns the direct and relative spatial features from the scene, and GAN is proposed to constrain the depth estimation generator with multiple discriminators.

\subsection{Self-supervised Learning Methods}

Among all the depth estimation techniques that have been proposed, depth-supervised methods are classically the most explored ones as they are intuitive and prone to implementation.
However, depth-supervised methods require numerous images with annotated ground-truth depth maps during training.
Hence, researchers explore self-supervised methods for monocular depth estimation without ground-truth depth maps. 
Self-supervised depth estimation models are typically trained on stereo images or continuous frame sequences constrained by stereo geometry relations.

Self-supervised learning depth estimation methods are developed from classic stereo matching methods, which obtain depths from left-right image pairs or adjacent frames.
In 2016, Garg \textit{et~al.} \citep{garg2016unsupervised} introduced a general framework to estimate monocular depth maps in an unsupervised manner with constraints on reconstruction loss, which measures the error between the re-projection image and the source image.
The general framework guided the re-projection images from the right-view image and the estimated depth that was estimated from the input left-view image with a deep CNN.
Godard \textit{et~al.} introduced Monodepth \citep{godard2017unsupervised} to reconstruct the left-right consistent depths, which are constrained by the reconstruction loss, the disparity smoothness loss, and the left-right disparity consistency.
Experimental results demonstrated that the multiple plausible error combination improved the prediction accuracy estimated from paired images.
Furthermore, Godard \textit{et~al.} \citep{godard2019digging} proposed Monodepth2, an auto-masking method for dynamic objects in lane scenes that minimized the multi-scale appearance matching loss and depth artifacts. This is the first self-supervised framework for inter-frame mechanisms, which greatly improves the applicability of depth estimation to various scenarios.
Inspired by the direct visual odometry, Wang \textit{et~al.} \citep{wang2018learning} argued that the model can be trained without the pose predictor. 
They also proved the feasibility of Differentiable Direct Visual Odometry (DDVO) with a depth normalization strategy.
As the re-projection error focus only on local patch similarity, Li \textit{et~al.} \citep{li2021adv} exploited global distribution differences though an adversarial loss.

For monocular depth estimation, self-supervised learning methods extract depth information directly from explicit geometric relations, including two primary manners: stereo matching and adjacent frames supervision.
In comparison to supervised learning methods, self-supervised learning methods do not require ground-truth depth maps, achieving high interpretability at the expense of lower accuracy.

\subsection{Reflective scene depth estimation}

In water scene depth estimation, it essentially pertains to the depth estimation of reflective scenes. The initial work was proposed by Yang \textit{et~al.} \citep{yang2015depth}, with a primary focus on matching symmetric images. On one hand, Yang's work demonstrated the feasibility of inferring depth from a single water scene image. On the other hand, their work was preliminary, lacking a standard dataset to evaluate the depth estimation results.

Further, Kawahara \textit{et~al.} \citep{kawahara2020appearance} presented a method for reconstructing 3D scene structure and high-dynamic range appearance from single-image water reflection photography, overcoming challenges posed by environmental illumination and water surface waves. It demonstrated that such images, combining direct and reflected scenes, enable self-calibrating HDR catadioptric stereo cameras, offering a novel approach to scene reconstruction and camera calibration.
However, it is noteworthy that above geometric-based methods was are low-precision and have harsh application scenarios.

\begin{figure*}[ht]
	\centering
	\includegraphics[width=\linewidth]{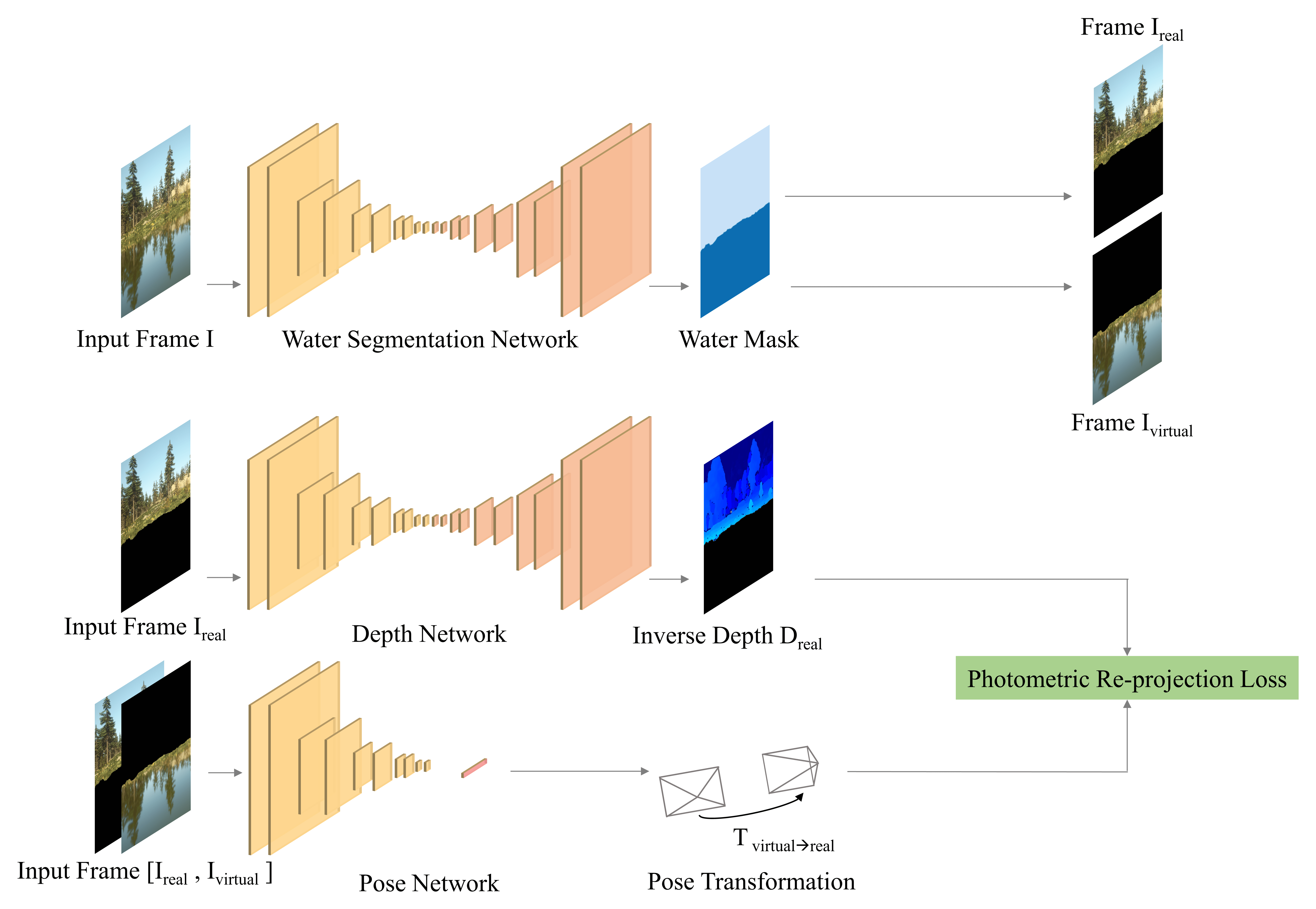}
	\caption{Complete framework of the proposed depth estimation method. The water segmentation network employs a standard convolutional U-Net for water area prediction. Similarly, the depth network employs a standard U-Net for depth prediction.The independent pose network predicts perspectives between real and virtual images.}
	\label{fig:netarc}
\end{figure*}

\section{Methodology}

The proposed two-stage monocular depth estimation on water scenes via specular reflection priors is shown in Figure \ref{fig:netarc}, where the inverse depth is estimated by minimizing the photometric re-projection error between the camera and corresponding mirrored virtual perspective in water reflection scenes.
To utilize the prior information of the reflective component, the first step is to capture the reflection image and the second step is to match the reflection image features with real scenes for depth estimation.
Therefore, the reflection prior-based self-supervised framework comprises two parts, the water surface segmentation module and the reflection image matching module.
For the water segmentation, the standard end-to-end U-Net \citep{ronneberger2015u}, which can be expanded in follow-up research, is employed to segment real areas and reflective components.
With the photometric re-projection constraint, the reflection image matching module works to simultaneously estimate the depth maps and camera pose transformation.
To match the reflections and source patterns, the photometric adaptive SSIM is introduced to the re-projection error to emphasize contrast and structural differences.
For reflection relations, real and virtual camera poses determine the water plane, whose depth can be filled in the segmented surface area.

\begin{algorithm}[tbp]
	\SetAlgoLined
	\KwData{$S(x)$:Segmentation network; $D(x)$:Depth network; $P(x,y)$:Pose network; $I$: Input frame;}
	\KwResult{Depth: $D$}
	
	Initial weight of $D(x)$ and $P(x)$\;
	
	Segment the water surface:\;
	\quad  \quad     $I_{real},I_{virtual}=S(I)$\;
	\While{not converged}{
		Estimate the pose transformation:\;
		\quad  \quad $T_{real\rightarrow virtual}=P(I_{real},I_{virtual})$\;
		Estimate the inverse depth: \;
		\quad  \quad $D_{real}=D(I_{real})$\;
		Re-project the virtual pixels into real pixels\;
		\quad  \quad $I_{virtual \rightarrow real}=I_{virtual}\left\langle {proj}\left(D_{real}, T_{real \rightarrow virtual}, K\right)\right\rangle$\;
		Compute the photometric re-projection error:\;
		\quad  \quad $L_{p}=\sum_{virtual} pe\left(I_{real}, I_{virtual \rightarrow real}\right)$\;
		Update the depth network by adaptive moment estimation optimizer\;
		\quad  \quad $\theta_D:=\mathop{argmin}_{\theta_D}{L_p}$\;
		Update the pose network by stochastic gradient descent optimizer\;
		\quad  \quad $\theta_P:=\mathop{argmin}_{\theta_P}{L_p}$;
	}	
	Estimate the water plane\;
	\quad  $\quad Plane:\leftarrow(T_{real->virtual})$\;
	Complete the depth of virtual pixels\;
	\quad  \quad $D=D_{real}+D_{virtual}$\;
	\caption{Reflection-supervised depth estimation}
	\label{al1}
\end{algorithm}

Algorithm \ref{al1} explains the reflection-supervised depth estimation. 
As shown in Algorithm \ref{al1}, the depth network is optimized by adaptive moment estimation optimizer and the pose network is optimized by stochastic gradient descent optimizer.

\subsection{Water Segmentation}

To segment the real scene and the virtual image reflected from the water surface, a standard end-to-end U-Net \citep{ronneberger2015u} is constructed to fit the water segmentation task in the proposed work, as shown in Figure \ref{fig:netarc}.
Considering the limited accuracy of the water segmentation module's baseline, our objective is to generate acceptable inputs for the subsequent depth estimation module.
Consequently, the lightweight backbone not only proves the feasibility of the proposed method, but also can be conveniently replaced and expanded in follow-up researches.

In the proposed multi-stage task, the initial water segmentation step is crucial as error accumulates.
A binary Dice loss \citep{milletari2016v}, which aims to overcome the data imbalance problem, is used as the water segmentation loss.
The Dice coefficient $D$, a measure used to assess sample similarity between two binary segmentation images, can be written as:

\begin{equation}
D=\frac{2 \sum p_{i} g_{i}}{\sum p_{i}^{2}+\sum g_{i}^{2}}
\end{equation}
where the sums traverse all pixels, of the estimated binary segmentation pixel $p_i$ and the ground-truth binary pixel $g_i$.
The Dice loss $\mathcal{L}_{Dice}$ can be stated as:

\begin{equation}
\mathcal{L}_{Dice}=1-\frac{2 \sum p_{i} g_{i}+\epsilon }{\sum p_{i}^{2}+\sum g_{i}^{2}+\epsilon }
\end{equation}
where the small positive value $\epsilon$ is used to prevent the denominator from being zero and smoothing gradients in optimization.
The Dice loss is a region-related loss, where the pixel loss is not only related to the predicted pixel, but also to the neighboring points.
Regardless of the image size, fixed size positive sample regions have the same loss. Thus the Dice loss allows the model to focus on negative samples.
In summary, the complexity of the water segmentation network is maintained at a low level, and the loss function is modified to construct a lightweight system.

\subsection{Self-supervised Depth Estimation}

Monocular depth estimation via photometric re-projection error reformulates the ill-posed depth estimation task as an interpretable multi-view synthesis problem.
In contrast to the re-projection loss typically adapted for adjacent frame supervision in previous models, the loss is applied to penalize the pixel projection error between the real and virtual camera pose in a single frame.

Based on the re-projection mechanism of Monodepth2 \citep{godard2019digging}, a PoseNet based on ResNet18 \citep{he2016deep} is constructed to predict the camera pose transformation matrix $T_{t->t^{\prime}}$ between the target image $I_{t^{\prime}}$ and the source image $I_{t}$, which is presented as below:

\begin{equation}
\begin{aligned}
T_{t->t^{\prime}}=PoseNet\left(I_{t}, I_{t^{\prime}}\right)
\end{aligned}
\end{equation}

The depth map $D_t$ is estimated through minimizes the photometric re-projection error $L_p$, where

\begin{equation}
\begin{aligned}
\label{deqn_ex1a}
L_{p} &=\sum_{t^{\prime}} p e\left(I_{t}, I_{t^{\prime} \rightarrow t}\right), \\
\text { and } \quad I_{t^{\prime} \rightarrow t} &=I_{t^{\prime}}\left\langle {proj}\left(D_{t}, T_{t \rightarrow t^{\prime}}, K\right)\right\rangle
\end{aligned}
\end{equation}
where $K$ are the camera intrinsics which are identical in all perspectives, $proj$ represents the 2D coordinates of target image from projected depths, $\left\langle \right\rangle$ represents a bilinear sampling operation, and $I_{t^{\prime} \rightarrow t}$ represents the pixel set that is reprojected from the original frame to the target frame.
$pe$ is the pixel photometric re-projection error incorporating SmoothL1 \citep{girshick2015fast} and Photometric Adaptive SSIM (PASSIM), which is developed from SSIM \citep{wang2004image}. Hence, $pe$ can be formulated as:

\begin{equation}
\begin{aligned}
p e\left(I_{t}, I_{t->t^{\prime}}\right)  = \frac{\alpha}{2}\left(1-\operatorname{PASSIM}\left(I_{t}, I_{t->t^{\prime}}\right)\right)^{\frac{1}{2} }+(1-\alpha)\left\|I_{t}-I_{t->t^{\prime}}\right\|_{1}
\end{aligned}
\end{equation}
where $\alpha = 0.75$. 
Since the value of PASSIM is commonly approximated to 1, the structural error is considered small and requires a square root operation.
As the image color distribution shown in Figure \ref{fig:refdemo}, it is worth noting that the luminance component of the inverted image on the reflective material is degraded against the source one, as evidenced by contrast is decayed in proportion to the luminance.
In PASSIM, we remove luminance comparison measurements and emphasize contrast comparison $c$ and structure comparison $s$, where photometric adaptive contrast is introduced. Specifically, PASSIM adapted for surface reflection is defined as:

\begin{figure}[tbp]
\centering
\includegraphics[width=\linewidth]{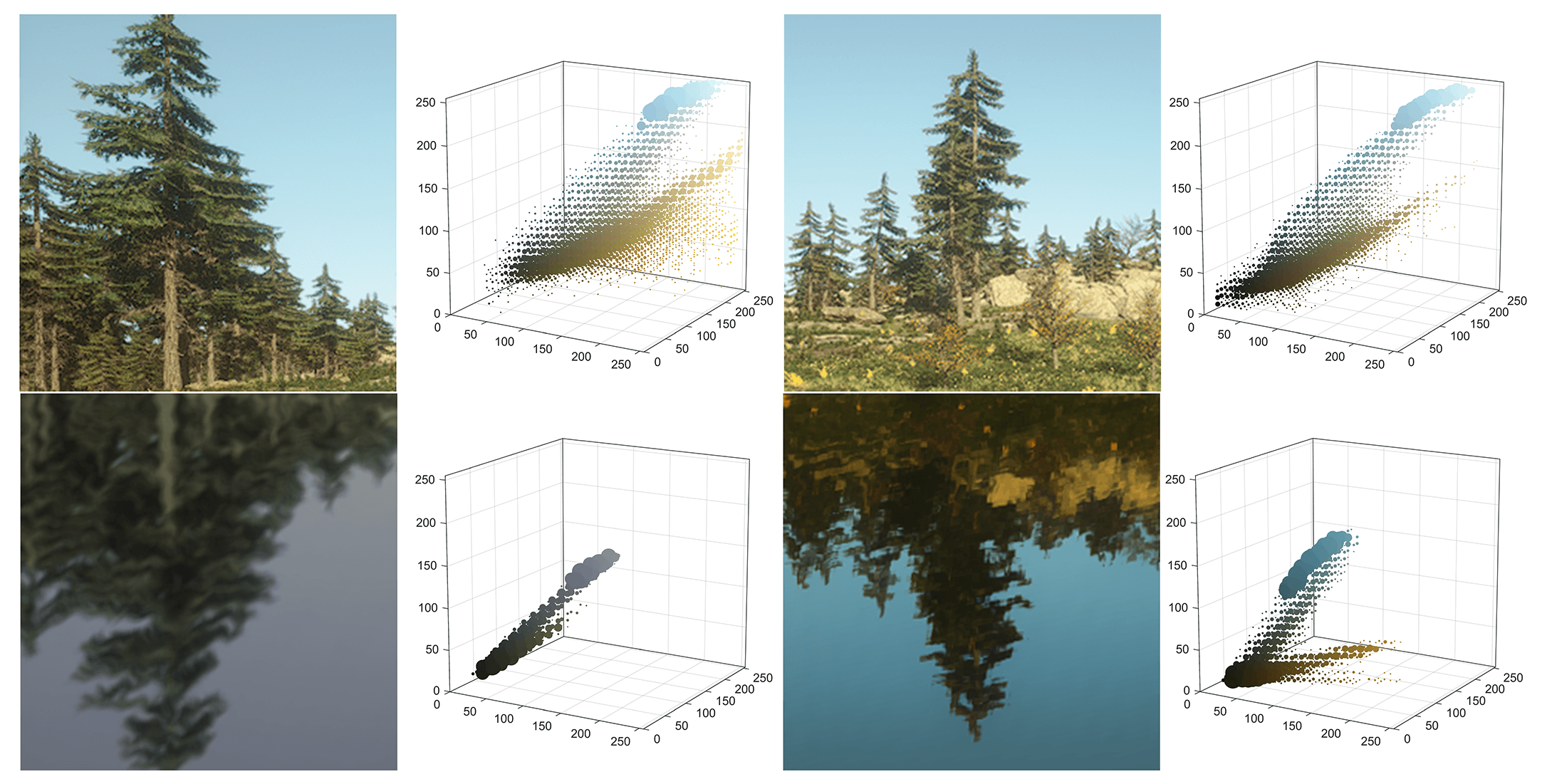}
\caption{Colour distribution of inverted and source image.
It can noticed that the distribution of inverted images is roughly proportionally decreased versus the source one.}
\label{fig:refdemo}
\end{figure}

\begin{equation}
\begin{aligned}
\operatorname{PASSIM}(\mathbf{x}, \mathbf{y})=[c(\mathbf{x}, \mathbf{y})]^{\beta}[s(\mathbf{x}, \mathbf{y})]^{\gamma}
\end{aligned}
\end{equation}
where

\begin{equation}
\begin{aligned}
c(\mathbf{x}, \mathbf{y})&=\frac{2 \hat{\sigma}_{x} \hat{\sigma}_{y}}{\hat{\sigma}_{x}^{2}+\hat{\sigma}_{y}^{2}}\\
s(\mathbf{x}, \mathbf{y})&=\frac{\hat{\sigma}_{x y}}{\hat{\sigma}_{x} \hat{\sigma}_{y}}\\
\hat{\sigma_{x}}=\frac{\sigma_{x}}{\mu_{x}},\enspace \hat{\sigma_{y}}&=\frac{\sigma_{y}}{\mu_{y}},\enspace \hat{\sigma}_{xy}=\frac{\sigma_{xy}}{\mu_{x}\mu_{y}}
\end{aligned}
\end{equation}
where $\mu _{x}$ is the average of $x$, $\mu _{y}$ is the average of $y$, $\sigma _{x}^{2}$ the variance of $x$, $\sigma _{y}^{2}$ the variance of $y$, 
and $\sigma_{xy}$ the covariance of $x$ and $y$.

For the extensive adaptability, the $\operatorname{PASSIM}$ is simplified as follows:

\begin{equation}
\begin{aligned}
\operatorname{PASSIM}(\mathbf{x}, \mathbf{y})=\frac{\mu_x \mu_y \sigma_{xy}+\epsilon}{\sigma_x^2 \mu_y^2+\sigma_y^2 \mu_x^2+\epsilon } 
\end{aligned}
\end{equation}
where small positive values $\epsilon$ prevent the denominator being zero and smooth the gradient in the optimization.

Same as Monodepth \citep{godard2017unsupervised}, an optional edge-aware smoothness is applied to smooth the estimated depth, which can be denoted as:

\begin{equation}
\begin{aligned}
L_{s} & = \left|\partial_{x} d_{t}^{*}\right| e^{-\left|\partial_{x} I_{t}\right|}+\left|\partial_{y} d_{t}^{*}\right| e^{-\left|\partial_{y} I_{t}\right|}
\end{aligned}
\end{equation}

Contrary to the intuitive perception in pose network design that clear reflection patterns are required for accurate pixel-level feature matching, this module performs the pose estimation with the raw blurred reflection patterns as it preserves the most image structure.

\subsection{Plane Depth Complement}

In addition, the depth maps and camera poses are optimized with photometric re-projection error.
Meanwhile, the depth and pose networks are well-trained to provide reliable camera poses and depth estimations, respectively.
According to the geometric constraints in the reflection scene, it is theoretically feasible to determine the reflection plane from the real and virtual camera poses to refine the depth of the water surface area. Hence, the reflection plane problem can be formulated as:

\begin{equation}
\begin{aligned}
distance(P,P_{real})=distance(P,P_{virtual})
\end{aligned}
\end{equation}
where $P_{real}$ is the real camera position and $P_{virtual}$ is the virtual one. 

Therefore this plane can be determined because $P_{real}$ and $P_{virtual}$ are known.
Next, the plane's projection is mapped to the estimated water surface area from the segmentation stage to obtain the depth relative to the real camera.
Theoretically, the geometric determination of water surface depth has unsurpassed accuracy and interpretability for reliable geometric constraints.

\section{Experiments}

\subsection{Experimental setup}

In this section, the proposed self-supervised depth estimation via specular reflection is validated upon the WRS dataset, manually generated from Unreal Engine 4. 
Notably, the dataset covers numerous challenging scenarios for depth estimation, including imbalanced lightings, occlusions, image halos and multiple ripple corrugations.
To determine the scale factor, we apply per-image median ground truth scaling \citep{zhou2017unsupervised} to restore scale information.
To evaluate the proposed model, we restrict the scene depths to a fixed distance between 0m and 120m and compare the performance with existing methods by widely-used evaluation metrics: absolute relative error (AbsRel), square relative error (SqRel), root mean square error (RMS), root mean square logarithmic error (RMS(${log}$)) \citep{karsch2014depth} and Accuracies of three different thresholds ($1.25$, $1.25^{2}$, $1.25^{3}$) which are shown as follows:

\begin{itemize}
\item{AbsRel$=\frac{1}{N}\textstyle \sum _{i\in N}\left |\frac{d_{i}-d_{i}^*}{d_{i}^*} \right |$}
\item{SqRel$=\frac{1}{N}\textstyle \sum _{i\in N}\frac{\left \|d_{i}-d_{i}^* \right \|_2}{d_{i}^*}$}
\item{RMS$=\sqrt{\frac{1}{N}\textstyle\sum_{i\in N} \left \| d_i-d_i^* \right \|_{2}}$}
\item{RMS$({log})=\sqrt{\frac{1}{N}\textstyle\sum_{i\in N} \left \| log(d_i)-log(d_i^*) \right \|_{2}}$}
\item{Accuracies$=max(\frac{d_{i}}{d_{i}^*},\frac{d_i^*}{d_{i}})=\delta< threshold$}
\end{itemize}

where $N$ donates the number of ground-truth depth pixels, $d_i$ donates the predicted depth value at pixel $i$ and $d_i^{*}$ is the ground-truth depth at pixel $i$.
Additionally, $threshold$ limits the correct percentage of pixels in the estimated depth, which can be taken as $1.25$, $1.25^{2}$, $1.25^{3}$.

\subsubsection{Water Reflection Scene Dataset}

\begin{figure}[tbp]
\centering
\includegraphics[width=0.8\linewidth]{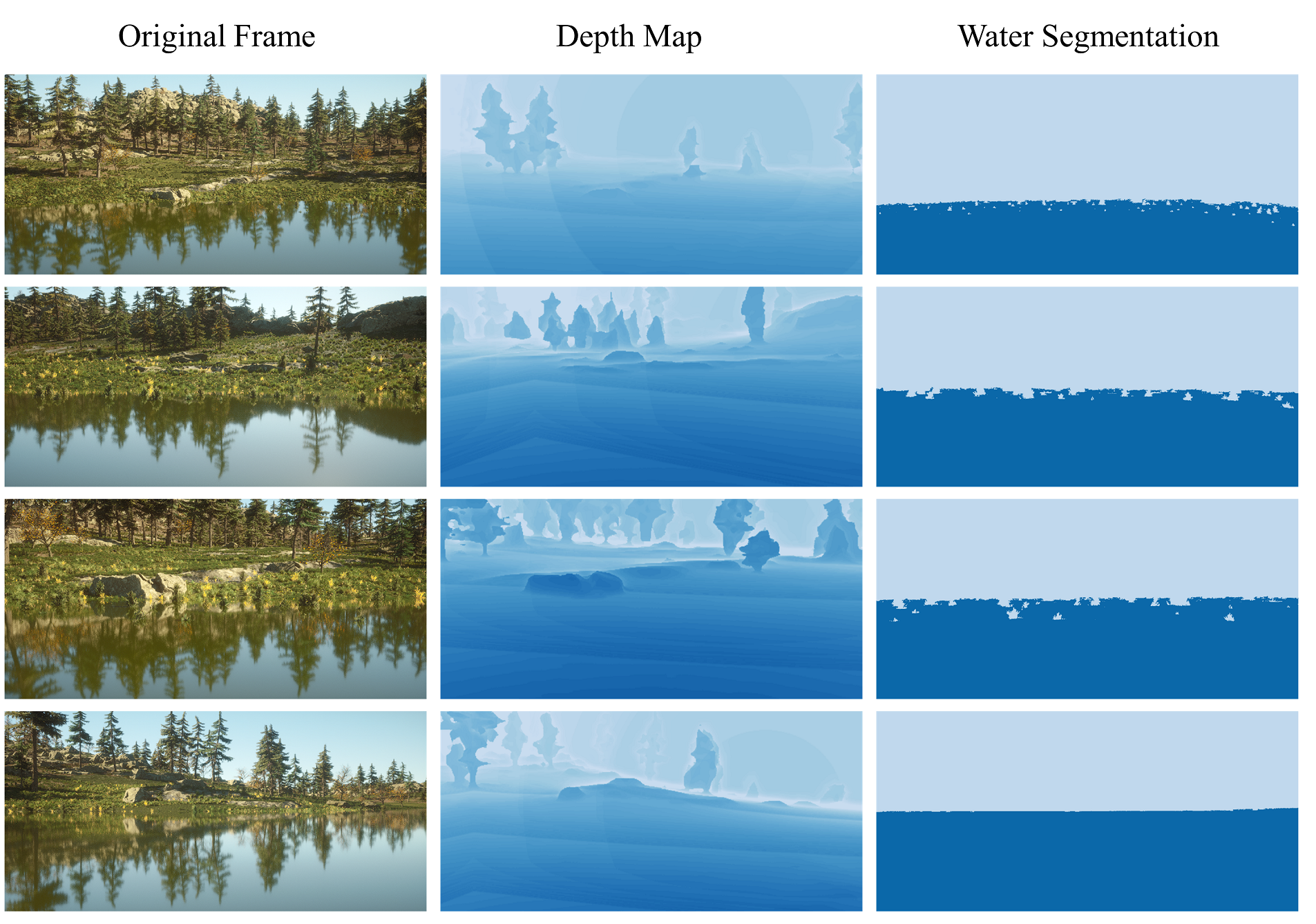}
\caption{Examples in the WRS dataset. 
We can observe that the depths and segmentataions are visually highly accurate.}
\label{fig:wrsdemo}
\end{figure}

The absence of a Water Reflection Scene dataset (WRS) limits the research of specular reflection problems in computer vision. 
To develop this research field, we create a large-scale specular reflection dataset for 3 disparate water scenes, including 3251 RGB images and depth pairs rendered from Unreal Engine 4.
In the WRS dataset, the original images and corresponding depths are 1200$\times$572 pixels with a few missing depth pixels on the boundaries due to the sampling problem in the virtual scene.
The WRS dataset collects most images from flowing water conditions with various complex light conditions to construct realistic outdoor scenes with clear water reflections, as shown in Figure \ref{fig:wrsdemo}.

In contrast to ground-truth depths from LiDAR sensors or generated from SFM methods, the depths generated from virtual reality scenes are more correct and eliminate laborious pre-processing.
On the other hand, the reflection dataset is limited to the ideal optical condition and rendering defects in virtual scenes.
As a remedy, the proposed method will be validated on real-world scenarios that are not included in the WRS dataset to prove its feasibility.
\textcolor{blue}{Further, The WRS dataset can now be accessed on \url{https://github.com/Mnster00/ReflectionDepth}.}

\subsubsection{Implement Details}
The system is implemented in PyTorch with a single Nvidia RTX 2080 Super 11G. 
To facilitate re-implementation, a standard 5-layer U-Net \citep{ronneberger2015u} is adopted as the backbone for the water segmentation and depth estimation.
As in Monodepth2 \citep{godard2019digging}, the ResNet18 weights on ImageNet \citep{russakovsky2015imagenet} as an initialization of PoseNet. 
The model is trained on the WRS dataset, whose homogenous scenarios lead to over-fitting potentially.
Therefore the pose network and the depth estimation network are pre-trained on the KITTI 2015 dataset \citep{geiger2012we}, and the water segmentation network is pre-trained on USVInland dataset \citep{cheng2021we}.
Pre-training is employed to expedite the convergence of training, but it does not impact the final accuracy achieved during the training process.
The same intrinsics are determined for testing images, setting the principal point to the image center and the focal length same as KITTI dataset \citep{geiger2012we}.
To obtain statistically significant results, we conduct depth estimation experiments on each model and compute the average values.

In the WRS dataset, the original images are 1200$\times$572 pixels and are resized to 1200$\times$600 pixels while removing the boundaries where the depth values do not exist.
None of the data augmentation strategies is applied to increase the scale of data.
To divide the training and testing split, we select 2600 unique RGB images and corresponding depths for training and 651 for testing without cross-validation.
In the pre-processing of size, the input image is down-sampled to 1024$\times$512 pixels, and the output is up-sampled to the original size by Bilinear interpolation to fit the network.

\subsection{Water Segmentation Results}

Water segmentation is the first stage of the depth estimation model, and its segmentation results are critical to the integrated model performance, as errors accumulate.
Therefore, following experiments are conducted to select a suitable network model that minimizes the attenuation of the depth estimation model's performance based on the model-predicted water segmentation results.
Table \ref{tab:waterablation} evaluates the water segmentation results estimated by various network structures, followed by the depth accuracy resulting from the depth estimation network.

\begin{table*}[htbp]
\centering
\caption{Depth accuracy comparison with different water segmentation results on the WRS dataset}
\resizebox{\linewidth}{!}{
\begin{tabular}{cccccccccccccc}
\toprule[1pt]
\multirow{2}{*}{Method}                           & \multicolumn{5}{c}{Water Segmentation Errors}                    	&    & \multicolumn{7}{c}{Depth Estimation Errors} \\ \cline{2-6} \cline{8-14} 
\specialrule{0em}{0pt}{2pt}
                                                  & AvgPrec{[}\%{]} & PRE{[}\%{]} & REC{[}\%{]} & FPR{[}\%{]} & FNR{[}\%{]}& & AbsRel   & SqRel    & RMS     & RMS(${log}$)   & $\delta_1${[}\%{]} & $\delta_2${[}\%{]} 	& $\delta_3${[}\%{]}	\\
\specialrule{0em}{0pt}{0pt}
\midrule[1pt]
U-Net                   & 95.54           & 99.01       & 92.33       & 0.99        & 7.67       & & 0.1210   & 0.8370   & 4.430  & 0.2060         & 88.60   & 95.30  & 98.10       \\
DeepLabv3			  & 95.65           & 99.13       & 93.08       & 0.87        & 6.92       & & 0.1190   & 0.8330   & 4.390  & 0.2050       & 89.10   & 95.50  & 98.20   \\
PDCNet \citep{yang2023pixel}			 & 95.73           & 99.18       & 93.19       & 0.82        & 6.81       & & 0.1180   & 0.8150   & 4.330  & 0.2020       & 89.40   & 95.70  & 98.20   \\
GT                                                & -               & -           & -           & -           & -          & & 0.1020   & 0.7630   & 3.940  & 0.1850      & 93.10	 &  97.90 & 98.60 \\
\bottomrule[1pt]
\end{tabular}}
\label{tab:waterablation}
\end{table*}

In Table \ref{tab:waterablation}, DeepLabv3+ \citep{chen2018encoder}, the widely-used state-of-the-art model, has similar accuracy to the U-Net which is supervised by Dice loss, with much higher complexity. 
Although the depth accuracy from the model-predicted segmentations is not as accurate as the ground-truth segmentations, it is sufficient for the proposed two-stage depth estimation task.
The accuracy results of advanced water segmentation methods on the proposed dataset are notably high, and the differences are not significant.
Therefore, U-Net with Dice loss is an acceptable choice to be the backbone of the first stage.

Noteworthy, besides the fully-supervised water segmentation, we have attempted to perform a self-supervised implementation that switched to the re-projection loss function after pre-training convergence.
However, this modification led to severe model degradation and the over-fitting of the entire depth estimation due to the homogeneous constraint.
A feasible solution to avoid the model degradation is to construct a discriminator, a module in the generative adversarial strategy, to sustain the model complexity, but there is no significant accuracy improvement in further experiments. 
In this case, it is not worthwhile to obtain a fully self-supervised framework at the cost of model interpretability and certainty.

\subsection{Ablation Experiments}

In the classical methods \citep{godard2019digging}, the photometric re-projection error combines SSIM \citep{wang2004image} and SmoothL1 \citep{girshick2015fast}, which provides structural and pixel-level similarity in an intuitive manner. 
To match the reflected and original patterns, we propose an adaptive similarity metric that combines SmoothL1 and PASSIM which focuses on proportionally decreased luminance and contrast.

For the structural error implementation, it is preferable to apply local metrics rather than global ones.
First, most visual statistical features are highly spatially non-stationary.
Second, the dominant image distortions, spatial and frequency variation, may be observed from the local assessment.
Finally, localized similarity measurement delivers more information about image degradation.
Same as the implementation of SSIM, PASSIM is computed within the local kernel, a circular-symmetric Gaussian weighting function, which traverses pixel-by-pixel over the entire image.
The local error implementation is revised as follows:

\begin{equation}
\begin{aligned}
\mu_{x} &={\textstyle \sum w_{i} x_{i} }\\
\sigma_{x}& =\sqrt{ {\textstyle \sum w_{i}\left(x_{i}-\mu_{x}\right)^{2} }} \\
\sigma_{x y} &={\textstyle \sum w_{i}\left(x_{i}-\mu_{x}\right)\left(y_{i}-\mu_{y}\right)}
\end{aligned}
\end{equation}
where $w_i$ is the Gaussian weight. Therefore, the window size is a crucial hyper-parameter that requires to be determined.

\begin{table}[tbp]
\centering
\caption{Comparison of accuracy on different kernel sizes.}
\resizebox{0.6\linewidth}{!}{
\begin{tabular}{ccccccccccc}
\toprule[1pt]
\multirow{2}{*}{Metric} & \multicolumn{5}{c}{Kernels}                                              &  & \multicolumn{3}{c}{Errors}          \\ \cline{2-6} \cline{8-10}
\specialrule{0em}{0pt}{2pt} 
                        & 3            & 5            & 7            & 9            & 11           &  & AbsRel & SqRel & RMS  \\ 
\specialrule{0em}{0pt}{0pt}
\midrule[1pt]
PASSIM                  & $\checkmark$ &              &              &              &              &  & 0.1610     & 1.2790    & 5.760              \\
(w/ gaussian)           &              & $\checkmark$ &              &              &              &  & 0.1290     & 0.8540    & 4.715              \\
                        &              &              & $\checkmark$ &              &              &  & 0.1330     & 0.9190    & 4.860              \\
                        &              &              &              & $\checkmark$ &              &  & 0.1340     & 0.9350    & 4.930              \\
                        &              &              &              &              & $\checkmark$ &  & 0.1380     & 0.9980    & 5.100                \\ 
\specialrule{0em}{0pt}{2pt} 
\hline
\specialrule{0em}{0pt}{2pt} 
PASSIM                  & $\checkmark$ &              &              &              &              &  & 0.1580     & 1.1830    & 5.520       \\
(w/o gaussian)          &              & $\checkmark$ &              &              &              &  & 0.1250     & 0.7970    & 4.510               \\
                        &              &              & $\checkmark$ &              &              &  & 0.1270     & 0.8150    & 4.630             \\
                        &              &              &              & $\checkmark$ &              &  & 0.1300     & 0.8360    & 4.670        \\
                        &              &              &              &              & $\checkmark$ &  & 0.1340     & 0.9020    & 4.840              \\
\bottomrule[1pt]
\end{tabular}}
\label{tab:lossablationkernel}
\end{table}

In Table \ref{tab:lossablationkernel}, we evaluate PASSIM-supervised depth performance on different kernels.
As shown in Table \ref{tab:lossablationkernel}, depth results on PASSIM without Gaussian weight components perform better because of the center-free spatial stochasticity of water ripples. 
The kernel size is highly correlated with the pixel drift, namely the shift of the pixel position in reflections triggered by water wave vibrations.
From experimental results, the pixel drift in most reflections on the WRS dataset is less than 5 units. 
In the following experiments, SSIM and PASSIM kernels are set to 5 without Gaussian weights.

\begin{table}[tbp]
\centering
\caption{Ablation experiments on the WRS dataset}
\begin{tabular}{ccccc}
\toprule[1pt]
Re-projection Error	 	& AbsRel     & SqRel     & RMS  	& RMS(${log}$) \\
\specialrule{0em}{0pt}{0pt}
\midrule[1pt]
SmoothL1 						& 0.2050     & 1.6040    & 6.580     	& 0.2860    	\\
SSIM + SmoothL1					& 0.1680     & 1.4130    & 6.120    	& 0.2450		\\
PASSIM + SmoothL1 				& 0.1210     & 0.8370    & 4.430     & 0.2060     	\\
\bottomrule[1pt]
\end{tabular}
\label{tab:lossablation}
\end{table}

Table \ref{tab:lossablation} demonstrates the accuracy comparison of depth estimation for various combinations of structural and pixel-level error, which indicates the importance of each component in the photometric re-projection.
The weight $\alpha$ of photometric re-projection errors is set to $0.75$, and PASSIM is enhanced through a one-half power operation.
The comparison of the experimental results illustrates that the combination of PASSIM and SmoothL1 benefits the accuracy of the depth estimates, and PASSIM is particularly efficient for matching reflections and original patterns.

\subsection{Comparisons with State-of-the-art Methods}

For the fairness of the comparison, each depth estimation method has been pre-trained on the KITTI 2015 dataset \citep{geiger2012we}.
Existing methods have the same experimental training parameters as their open source codes.
It is worth noting that some self-supervised depth estimation methods, such as Monodepth \citep{godard2017unsupervised}, Monodepth2 \citep{godard2019digging}, and DDVO \citep{wang2018learning}, are not included in the comparison as their pose estimation modules are devised for inter-frames and can not port to the intra-frame task on water reflection scenes.

\begin{figure*}[tbp]
\centering
\includegraphics[width=\linewidth]{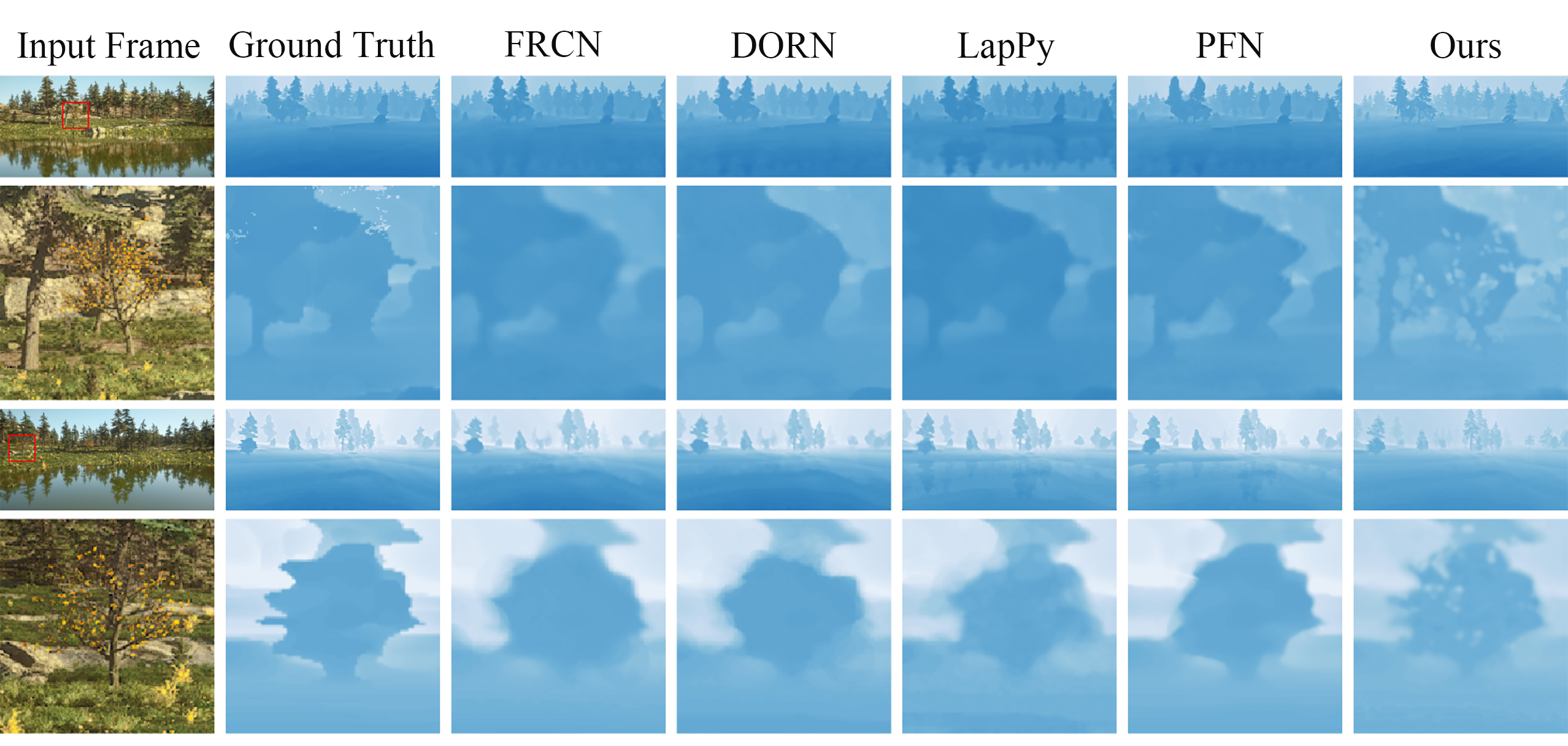}
\caption{Qualitative results on the WRS dataset. 
Our self-supervised method produces the sharpest depth maps, even outperforming the ground-truth results in visual details.}
\label{fig:sotaqualitative}
\end{figure*}

Figure \ref{fig:sotaqualitative} demonstrates the qualitative results on the WRS dataset. 
Rather than existing end-to-end methods being deceived by reflections, our method estimates depth maps that are not only visually accurate but also reconstruct a smooth water surface resulting from camera poses.
Note that the proposed method provides even more object details than the ground-truth depths.
Object meshes are monolithic planes in virtual scenes, which produces sharp boundaries in the rendered image by high-quality textures.
As a result, pixels in the same mesh have the same depth, which leads to insufficient details in the ground-truth maps.
In contrary to vague depth supervision, the self-supervised solution matches the reflection priors and source patterns directly within the same frame, which is strictly correct at the pixel level.

Since the determined plane reconstructs the water surface, it is unfair to count the water area result in the total error for other existing methods directly.
Therefore, quantitative comparisons are provided in two manners: the entire area and the area without the reflective component, as shown in Table \ref{tab:fullsota} and \ref{tab:removesota}.

\begin{table}[tbp]
\centering
\caption{Entire depth comparison between our method and the state-of-the-art methods on the WRS dataset}
\begin{tabular}{cccccccccc}
\toprule[1pt]
\multirow{2}{*}{Method}                    & \multicolumn{3}{c}{Errors}             &    & \multicolumn{3}{c}{Accuracies} \\ \cline{2-4} \cline{6-8} 
\specialrule{0em}{0pt}{2pt}
						   & AbsRel   	& SqRel   & RMS  	& & $\delta_1${[}\%{]} & $\delta_2${[}\%{]} 	& $\delta_3${[}\%{]}	 \\ 
\specialrule{0em}{0pt}{0pt}
\midrule[1pt]
FCRN \citep{laina2016deeper}            	& 0.1910     & 1.5980    & 6.690  &   & 71.90        & 87.15          & 97.00           \\
BTS \citep{zhang2018progressive}     		& 0.1890     & 1.5780    & 5.640   &    & 72.80         & 87.40           & 96.90            \\
DORN \citep{fu2018deep}                 	& 0.1730     & 1.4160    & 6.250   &    & 76.00        & 89.10         & 97.30               \\
LapPy \citep{song2021monocular} 			& 0.1820     & 1.4950    & 6.420     &    & 73.90  		& 88.20  		& 97.10                         \\
AdaBins \citep{bhat2021adabins} 			& 0.1730     & 1.4180    & 6.250   &    & 75.90 		& 89.10 		& \textcolor{blue}{97.40}                         \\
PFN \citep{lu2022pyramid}             		& 0.1760     & 1.4360    & 6.340   &    &  75.30        & 88.90         	& 97.20              \\ 
NeWCRFs \citep{yuan2022new}   	& \textcolor{blue}{0.1700}     & \textcolor{blue}{1.4030}    & \textcolor{blue}{6.180}   &    &  \textcolor{blue}{76.90}        & \textcolor{blue}{89.90}         	& \textcolor{blue}{97.40}              \\
Ours                                		& \textcolor{red}{0.1210}     & \textcolor{red}{0.8370}    & \textcolor{red}{4.430}    &	   & \textcolor{red}{88.60}   		& \textcolor{red}{95.30}  		& \textcolor{red}{98.10}                           \\ 
\bottomrule[1pt]
\end{tabular}
\label{tab:fullsota}
\end{table}

Quantitative evaluation results for entire image are provided in Table \ref{tab:fullsota}, where the proposed model outperforms other existing methods in six metrics. 
Compared to NeWCRFs, the proposed model reduces 28.82\% error on AbsRel, 40.34\% error on SqRel and 28.32\% error on RMS.
Furthermore, the accuracy advantage of the proposed method comes not only from the accurate estimation of the water plane but also from the reliable reconstruction of the scene objects by reflection priors.

\begin{table}[tbp]
\centering
\caption{Non-reflective component depth comparison between our method and the state-of-the-art methods on the WRS dataset}
\begin{tabular}{cccccccccc}
\toprule[1pt]
\multirow{2}{*}{Method}                    & \multicolumn{3}{c}{Errors}                    	&    & \multicolumn{3}{c}{Accuracies} \\ \cline{2-4} \cline{6-8} 
\specialrule{0em}{0pt}{2pt}
						   & AbsRel   	& SqRel   & RMS  	&  & $\delta_1${[}\%{]} & $\delta_2${[}\%{]} 	& $\delta_3${[}\%{]}	 \\ 
\specialrule{0em}{0pt}{0pt}
\midrule[1pt]
FCRN \citep{laina2016deeper}          	& 0.1750     & 1.3990    & 6.070 	&    		& 73.30        	& 88.20          & 96.80               \\
BTS \citep{zhang2018progressive}     	& 0.1520     & 1.1780    & 5.510   	&  		& 79.00        	& 90.10          & 97.60            \\
DORN \citep{fu2018deep}                	& 0.1620     & 1.2820    & 5.780  	&     	& 76.70         & 88.70         & 97.10               \\
LapPy \citep{song2021monocular} 		& \textcolor{red}{0.1330}    & \textcolor{red}{0.9910}    & \textcolor{red}{5.000}    &    		& \textcolor{red}{84.50}  		& \textcolor{red}{93.50}		& \textcolor{red}{97.90}                         \\
AdaBins \citep{bhat2021adabins} 		& 0.1610     & 1.2920   & 5.760     &    		& 75.20  		& 90.60  		& 97.50                         \\
PFN \citep{lu2022pyramid}             	& 0.1580     & 1.2450   & 5.680  	&     	& 79.80         & 90.90         & 97.60             \\ 
NeWCRFs \citep{yuan2022new}  	& 0.1370     & 1.0210    & 5.090   &    &  82.30        & 93.30         	& \textcolor{blue}{97.70}              \\
Ours                                	& \textcolor{blue}{0.1360}     & \textcolor{blue}{0.9990}   & \textcolor{blue}{5.010}    &    		& \textcolor{blue}{84.20} 		& \textcolor{red}{93.50}  		& 97.60                         \\ 
\bottomrule[1pt]
\end{tabular}
\label{tab:removesota}
\end{table}

Table \ref{tab:removesota} demonstrates that the proposed method achieves the second best performance in the non-reflective areas. 
Compared to the best method LapPy\citep{song2021monocular}, the proposed method is close in each metrics, with less than 3.0\% error.
Compared to depth-supervised methods, self-supervision inherently exhibit weaker accuracy due to their non-rigid constraints. In other words, the self-supervision methods are unable to match all pixels, whereas depth-supervised methods can capture the depth information of every pixel.
However, significant improvements in various metrics that include the water surface confirm the feasibility of the self-supervised solution.

\begin{table}[tbp]
\centering
\caption{Comparison of the model size between our method and state-of-the-art methods}
\begin{tabular}{crr}
\toprule[1pt]
Method 							& Model Size  &  Running Time \\ 
\midrule[1pt]
FCRN \citep{laina2016deeper}               &    230.2M              	&	1015.260ms\\
BTS \citep{lee2019big}         			&       202.8M     				&	708.372ms\\
DORN \citep{fu2018deep}                  	&      126.1M             	&	273.989ms		\\
AdaBins \citep{bhat2021adabins}  			&    270.1M          		&		-	\\
PFN \citep{lu2022pyramid}            		&          570.2M         	&	1654.502ms      \\ 
LapPy \citep{song2021monocular} 	           &    73.5M          		&	161.087ms	\\ 
Ours                        			 &          46.2M            	&	78.308ms  		\\      
\bottomrule[1pt]
\end{tabular}
\label{tab:param}
\end{table}

We also compare the model size and single-frame running time between the proposed method with existing methods at an input image of 1024$\times$512 pixels.
As illustrated in Table \ref{tab:param}, the size of the proposed method is the smallest, while the second-smallest size of LapPy is 59\% larger.
Meanwhile, compared to existing methods, our approach has the shortest processing time per frame.
The PFN is the largest model but has lower accuracy than our method.
Note that our method achieves the best performance on the WRS dataset, compared to LapPy, which has inferior accuracy.

It reveals that the proposed method is efficacious for the reflection scene depth estimation task and can achieve superior performance with the base backbone.
According to qualitative and quantitative results, the proposed method outperforms state-of-the-art monocular depth estimators in terms of the lower water surface estimation error and higher global accuracy.

\subsection{Real-world Scenario Results}

\begin{figure*}[h]
	\centering
	\includegraphics[width=\linewidth]{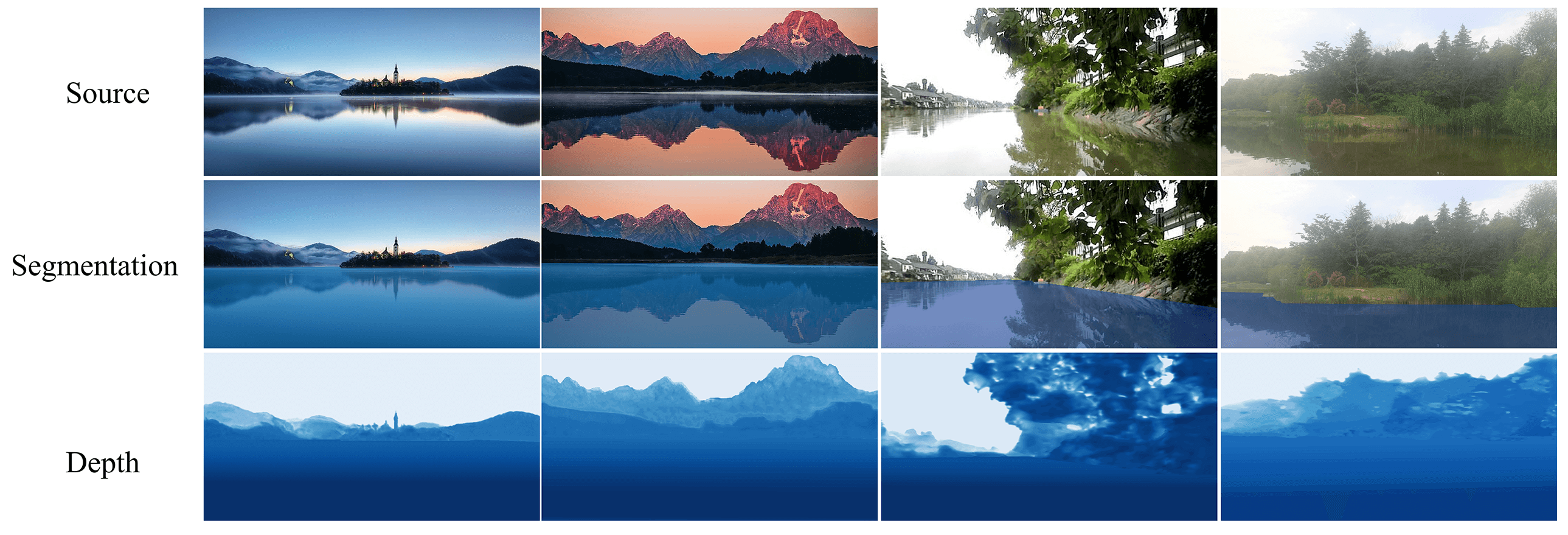}
	\caption{Depth estimation results from real-world scenarios. 
		For quantitative experiments, the real-world results are visually accurate.}
	\label{fig:realdemo}
\end{figure*} 

Real-world scenes with complex lighting conditions, variable shooting distances, heavy fog obscuration, and other uncertainties are considerable challenges for the proposed method.
Another problem is that our model is trained on a virtual dataset. Thus it is compulsory to be validated in real scenarios to prove its effectiveness in various scenarios.

A massive mixture of open copyright images from the internet and the WRS dataset form a large-scale dataset for real-world applications, including 611 web images and 3251 virtual images from the original dataset with specular reflections.
Benefiting from the self-supervised approach, numerous web images do not require laborious depth annotations.
Therefore, only the water segmentation results need to be noticed to ensure that the input for depth estimation is reliable.
Simultaneously, the kernel size of PASSIM is set to 11 to tackle the severe ripples with high pixel drift.

We show randomly selected web images with segmentations and depth results in Figure \ref{fig:realdemo} to prove the feasibility of the proposed method in a real-world depth estimation task.
Qualitative results are visually accurate in landscape scenes with reflections on calm water.
The depths behind the edges of the front mountains are not correct due to the lack of supervision of such occluded areas in the self-supervised training.
Conversely, the scene depths, in which reflections and real areas are simultaneously focused, are accurate because it is highly supervised.
Although the present method yields correct depth information in most real-world scenarios, it is limited to clear reflections on the water surface, which has primarily specular reflections and light surface ripples.

\section{Conclusion and future works}

In this paper, we provide a depth estimation framework for intra-frame information interaction in the presence of specular reflections.
The proposed self-supervised method, which is the first work to reformulate the single image depth estimation as a multi-view synthesis problem, yields state-of-the-art accuracy on both virtual and real-world water scenes.
Nevertheless, reflection-prior approaches require clear specular reflections and smooth surfaces.
Hence, the specular reflection problem in computer vision is not only applicable to water surface scenes, but various particular scenarios are worth exploring, such as multi-plane reflections and reflective irregular-surface objects.
The reflection information-based intra-frame supervision mechanism can greatly expand the wide applicability of depth estimation, as demonstrated in the possibility of applying unlabelled images on reflective scenes for self-supervised depth estimation methods.
The challenge in further research of deep learning-based solutions to specular reflection problems involves, on the one hand, the construction of reliable geometric constraints and, on the other hand, the creation of large-scale specular reflection datasets.

\printcredits

\bibliographystyle{model1-num-names}

\end{document}